\documentclass{article}

\usepackage{PRIMEarxiv}

\usepackage[utf8]{inputenc} 
\usepackage[T1]{fontenc}    
\usepackage{hyperref}       
\usepackage{url}            
\usepackage{booktabs}       
\usepackage{amsfonts}       
\usepackage{nicefrac}       
\usepackage{microtype}      
\usepackage{lipsum}
\usepackage{fancyhdr}       
\usepackage{graphicx}       
\usepackage{authblk}
\graphicspath{{media/}}     
\usepackage{bm}
\usepackage{subfigure}
\usepackage{here}

\title{Optimal normalization in quantum-classical hybrid models for anti-cancer drug response prediction
}

\author[,1]{%
Takafumi Ito\thanks{ito-takafumi0906@g.ecc.u-tokyo.ac.jp}\textsuperscript{\hspace{0.5em}}}
\author[,2]{%
Lysenko Artem\thanks{alysenko@g.ecc.u-tokyo.ac.jp}\textsuperscript{\hspace{0.5em}}}
\author[,1,2]{%
Tatsuhiko Tsunoda\thanks{tsunoda@bs.s.u-tokyo.ac.jp}\textsuperscript{\hspace{0.5em}}}
\affil[1]{Laboratory for Medical Science Mathematics, Department of Computational Biology and Medical Sciences,\newline Graduate School of Frontier Sciences, The University of Tokyo, Tokyo, Japan}
\affil[2]{Laboratory for Medical Science Mathematics, Department of Biological Sciences,\newline School of Science, The University of Tokyo, Tokyo, Japan}

\begin{document}
\maketitle

\begin{abstract}
Quantum-classical Hybrid Machine Learning (QHML) models are recognized for their robust performance and high generalization ability even for relatively small datasets. These qualities offer unique advantages for anti-cancer drug response prediction, where the number of available samples is typically small. However, such hybrid models appear to be very sensitive to the data encoding used at the interface of a neural network and a quantum circuit, with suboptimal choices leading to stability issues. To address this problem, we propose a novel strategy that uses a normalization function based on a moderated gradient version of the $\tanh$. This method transforms the outputs of the neural networks without concentrating them at the extreme value ranges. Our idea was evaluated on a dataset of gene expression and drug response measurements for various cancer cell lines, where we compared the prediction performance of a classical deep learning model and several QHML models. These results confirmed that QHML performed better than the classical models when data was optimally normalized. This study opens up new possibilities for biomedical data analysis using quantum computers.
\end{abstract}


\section{Background}
Due to extreme heterogeneity of cancers no therapies are universally effective, and therefore it is particularly important to select optimal anti-cancer drugs for each individual patient. Computational predictive models are becoming increasingly essential for streamlining such choices based on various evidence from previous studies. Among possible approaches, deep learning-based methods have a number of advantages over conventional machine learning due to their superior feature extraction ability \cite{sharifi2019moli, sharma2023deepinsight}. 
On the other hand, deep learning-based methods sometimes suffer from poor generalization ability due to the activation function, which is an important component for its nonlinearity. This is thought to occur because the gradient of the activation function changes abruptly when the input is close to zero, inappropriately amplifying small data fluctuations \cite{xie2020smooth}. This problem is especially significant when the dataset is small or high-dimensional, since the distances between data points are large, making it difficult to interpolate.
Previous studies revealed that gene expression data are highly valuable for anti-cancer drug response prediction \cite{geeleher2014clinical}. Gene expression data are high-dimensional data with about 20,000 dimensions, and they are too large compared to the number of samples that can be used for prediction ($\sim$1000). Therefore, deep learning models tend to lose their generalization ability. \\
Quantum-classical Hybrid Machine Learning (QHML) is one of the emerging solutions for this problem. Previous studies found that QHML performs better than the classical model in some cases  \cite{senokosov2024quantum}. QHML is a model that combines trainable quantum circuits and traditional neural networks to take advantage of both high generalization ability of quantum machine learning and feature extraction ability of deep learning. Because quantum machine learning models process input data based on smooth quantum operations, they are expected to successfully interpolate the gaps in a small number of high-dimensional data and achieve high generalization performance. One limitation is that quantum models currently struggle to directly handle high-dimensional data because of current hardware limitations, such as the limited number of qubits and short decoherence times. This can be effectively addressed by using a neural network to map the input data into some compact embeddings that are sufficiently small to accommodate.\\
These characteristics of QHML make this system appear attractive for the task of anti-cancer drug response prediction. However, after some trials, we found that the performance of the trained QHML model is unstable. Among the possible causes, we considered the information flow between the deep learning part and the Parameterized Quantum Circuit (PQC) as one of the key aspects. In the proposed architecture, the embeddings extracted by the neural network are passed to the quantum circuit using rotation angle encoding. Quantum rotation gates have a periodicity of $2\pi$, and several inputs could be considered identical, which may lead to unstable convergence during training. There is a study using $\tanh$ to normalize the input of the quantum circuit, but the gradients around zero are too steep, and many input values are mapped onto the most extreme upper and lower ranges of the scale. This can be mitigated by adjusting this normalization function by introducing a “gradual $\tanh$". This function eliminates the periodicity and concentration of the output values and thus enables a more stable training. To evaluate this approach, we have conducted five response prediction runs for different anti-cancer drugs and found that appropriate normalization enhances the performance of the quantum-classical hybrid model. 
We also investigated how the normalization function transforms values in the trained model, suggesting that the proposed normalization function may solve two problems that hinder learning stability: the periodicity of quantum gates and the crowding of values due to normalization. This study provides insight on how to pass values with care when combining deep learning models with quantum machine learning for anti-cancer drug response prediction.
\section{Methods}
\label{sec:headings}
In this section, we describe the detailed methods for anti-cancer drug response prediction with QHML. We used PyTorch \cite{paszke2019pytorch} and Pennylane \cite{bergholm2018pennylane} to implement the entire model.
\subsection{Quantum-classical hybrid machine learning model architecture}
\begin{figure}[h]
  \centering
  \includegraphics[scale=0.45, trim={1mm 1mm 1mm 1mm}]{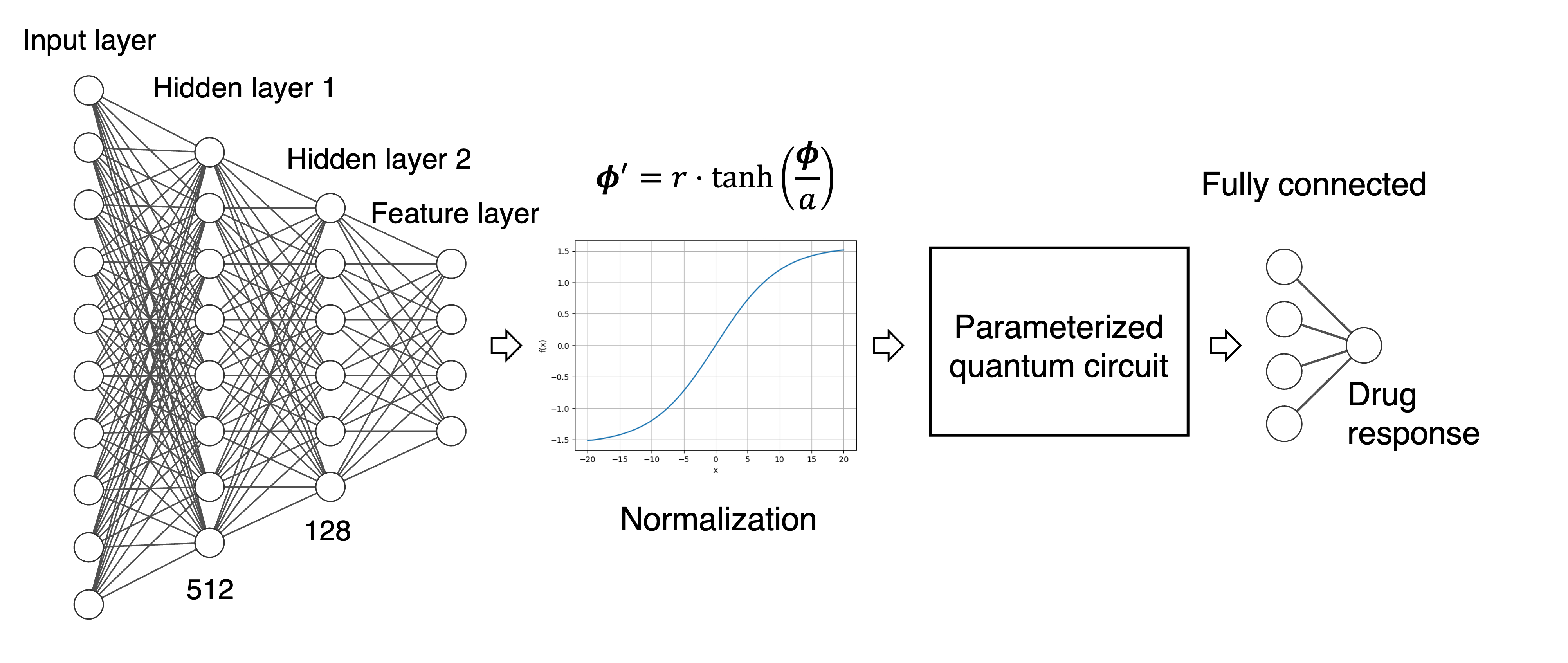}
  \hspace{-0.4cm}
  \caption{Overview of the quantum-classical hybrid machine learning model. Our proposed normalization function normalizes features effectively for inputting to the quantum circuits.}
  \label{fig:fig1}
\end{figure}

Here, we employed the quantum-classical hybrid model with three parts: a classical neural network encoder, a normalization function, and PQC (Figure \ref{fig:fig1}). In this study all quantum computation was done on a noiseless simulator, and the quantum states were calculated exactly.
\subsubsection{Classical neural network for data encoding}
The model first transforms input gene expression data into a low-dimensional embedding suitable for input to the quantum circuit by using a neural network encoder. Specifically, it is a three-layer neural network, with SiLU \cite{hendrycks2016gaussian} as the activation function. In addition, batch normalization is applied between the linear layer and the activation function. The dimensions of hidden layer 1 and hidden layer 2 are fixed to 512 and 128, while the final layer was adjusted to match the size of the subsequent quantum circuit.
\subsubsection{Normalization function}
In our model, the outputs of the encoder are used as angles of the initial rotation gates when initializing the circuit. The rotation gate has a periodicity of $2\pi$ and therefore multiple values represent the same angle of rotation, which can lead to unstable training. A previous study \cite{mari2020transfer} used a normalization function with tanh (\ref{eq:eq1}) in their implementation:
\begin{equation}
    \bm{\phi'} = \frac{\pi}{2}\tanh(\bm{\phi}).
    \label{eq:eq1}
\end{equation}
$\bm{\phi}$ is the output vector of the classical part, and $\bm{\phi'}$ is the converted vector that is passed to the quantum circuit. This function smoothly converts values between $-\frac{\pi}{2}$ and $\frac{\pi}{2}$ and thereby avoiding the periodicity problem. One of the disadvantages of using the equation \ref{eq:eq1} is that the gradient around zero is too steep, and many converted inputs will take values very close to $-\frac{\pi}{2}$ or $\frac{\pi}{2}$. This causes difficulty in training. The relationship between the $\tanh$ function and possible complications during training was already previously reported in the field of classical deep learning \cite{glorot2010understanding}. Therefore, we propose to use the following transformation. (\ref{eq:eq2}):
\begin{equation}
    \bm{\phi'} = r \cdot \tanh(\frac{\bm{\phi}}{a}).
    \label{eq:eq2}
\end{equation}
This function is a generalization of Equation 1 and smoothly transforms any real number into the range $-r\sim r$. By taking a sufficiently large value of $a$, we can make a function with a gentle gradient, which solves the problem of value crowding. We also consider the value of $r$ in this study. Although a large value of $r$ increases the range of possible values after normalization and may improve the expression ability, a periodicity problem arises when $r$ is greater than $\pi$. In addition, a small value of $r$ is expected to have a positive effect on regularization. 

\subsubsection{Parameterized quantum circuit}
\begin{figure}[H]
  \centering
  \includegraphics[width=0.90\columnwidth, trim={1mm 1mm 1mm 1mm}]{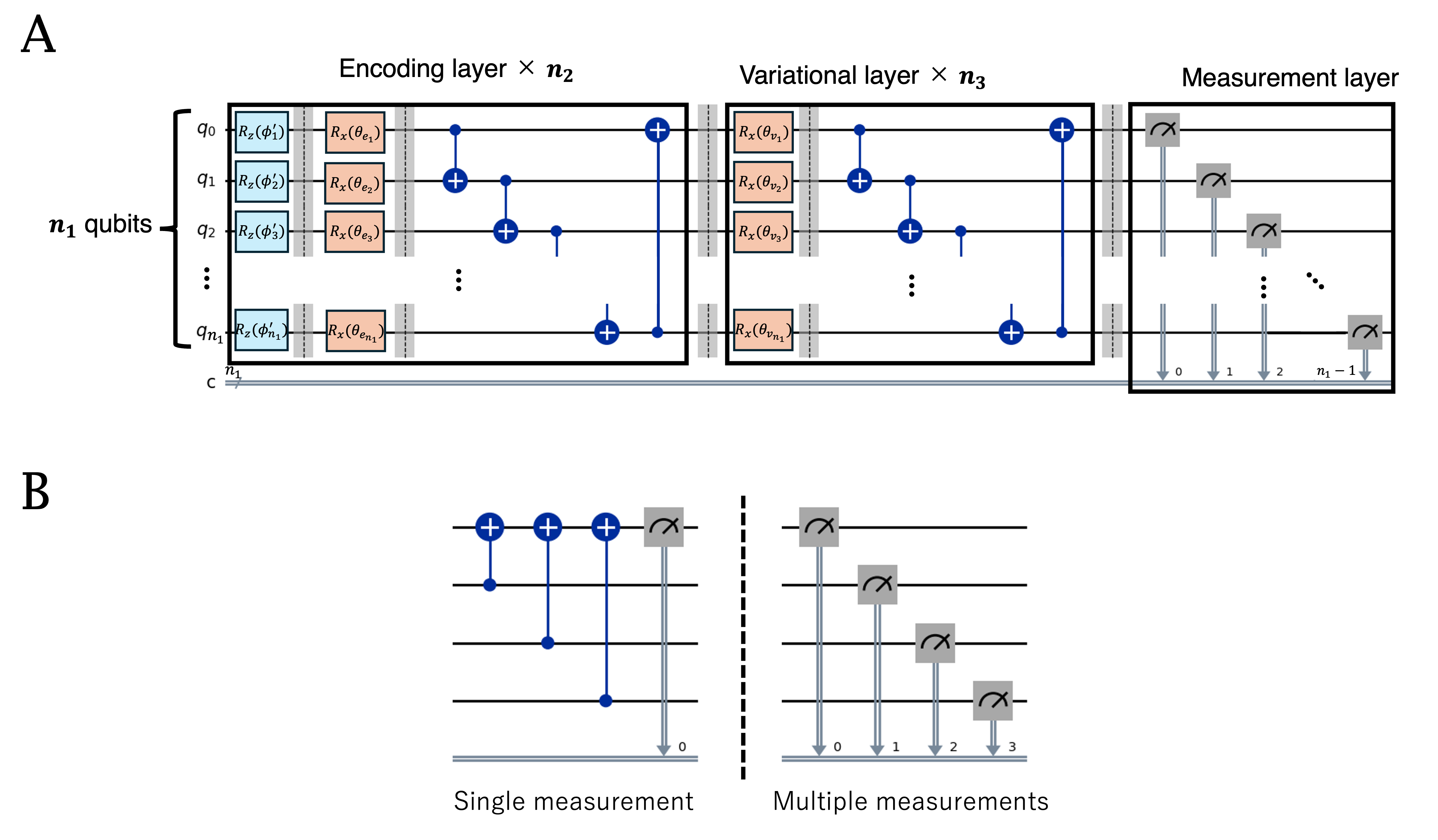}
  \caption{Parameterized quantum circuit architecture. (A) The parameterized quantum circuit consists of three parts and is characterized by 3 hyperparameters ($n_1$,$n_2$, and $n_3$). The measurement layer has two types described in Figure 2B. (B) We prepared two types of measurement layer. The difference is whether the integrations of values are done inside or outside of the quantum circuit.}
  \label{fig:fig2}
\end{figure}

After the normalization, the feature is passed to PQC. Figure \ref{fig:fig2}A shows the PQC design for our model. This circuit is almost the same as the one used in \cite{sagingalieva2023hybrid}. The blue gates are Z-rotation gates, and the rotation angles are determined by the input of the quantum circuit. The red gates are X-rotation gates, and the angles are determined by adjustable parameters. \\
The PQC is divided into three parts: the encoding layer, the variational layer, and the measurement layer. We used $n_1$ qubits for PQC. In the encoding layer, normalized features are passed to the PQC as quantum states. This encoding method is proposed in \cite{perez2020data} and \cite{schuld2021effect}. Repetition of the encoding layers $n_2$ times allows  for greater data encoding efficiently with a relatively small number of qubits. Next part is the variational layer, where parameterized gates change the quantum state. To achieve sufficient representation ability, this part is repeated $n_3$ times. The last part checks the state of the output qubits using a set of Pauli Z measurements. We prepared two types of measurement layers (Figure \ref{fig:fig2}B). When we used multiple measurements, we applied a single fully connected layer last to get scalar drug response values. The single measurements layer integrates values using CNOT gates and predicts  the drug response with a single measurement. All the calculations in this study were done on a noiseless simulator, so the expectation values are calculated exactly. We implemented all the quantum circuits with Pennylane \cite{bergholm2018pennylane}, which is a specialized library for quantum machine learning.
\subsection{Experimental setup}
We trained all the models with the same settings. We used mean squared error for the loss function and Adam\cite{kingma2014adam} for the optimizer. The batch size was 128, and we trained the model for 100 epochs.\\
For training and test data, we used cell line data obtained from the GDSC database. The GDSC database stores omics data on cell lines and their response values to various drugs \cite{yang2012genomics}. The data were used in a previous study \cite{sharifi2019moli}, and were obtained in the preprocessed form from the Zenodo repository maintained by the original authors. (https://zenodo.org/record/4036592) We randomly split the data 9:1 for training and test data. TGene expression data was the sole input and was already converted into TPM, log-transformed. In addition, standardized and homogenized to ensure compatibility between the data from different experiments \cite{geeleher2014clinical}. To exclude non-informative genes, we only used genes with training data variance above 0.1. For the drug response label, we have used continuous log(IC50) values for training and binary response labels for validation data. We used response data for five anti-cancer drugs with a sufficiently large sample size and relatively little bias in drug response: Cetuximab (Non-Responder(NR): 735 samples, Responder(R): 121 samples), Cisplatin (NR: 752 samples, R: 77 samples), Docetaxel (NR: 764 samples, R: 65 samples), Erlotinib (NR: 298 samples, R: 64 samples), and Gemcitabine (NR: 790 samples, R: 54 samples). We normalized log(IC50) values to $-1\sim1$ with equation \ref{eq:eq3} so that the value range of the quantum circuit output matches:
\begin{equation}
    DrugResponse_{norm} = \tanh (\frac{log(IC50) - mean(log(IC50))}{std(log(IC50))})
    \label{eq:eq3}
\end{equation}

\section{Results}
\subsection{Performance evaluation with cross validation}
We compared the prediction performance of the classical model, the proposed quantum-classical hybrid model, and quantum-classical hybrid models with different normalization methods. The classical model is the proposed model without the normalization and the quantum circuit, and with one additional linear layer instead. The quantum-classical hybrid model used for comparison consisted of three models: one without normalization (Equation \ref{eq:eq4}), one with normalization using nn.Layernorm from PyTorch (Equation \ref{eq:eq5}) (used in \cite{alam2021quantum}), and the last one with normalization using $\tanh$ (Equation \ref{eq:eq1}).
\begin{equation}
    \bm{\phi'} = \bm{\phi}
    \label{eq:eq4}
\end{equation}
\begin{equation}
    \bm{\phi'} = nn.Layernorm(\bm{\phi})
    \label{eq:eq5}
\end{equation}
We used multiple measurements for the measurement layer of these models. For the proposed method(Equation \ref{eq:eq2}), we prepared two models and employed each measurement layer. The hyperparameters $a$ and $r$ of the proposed method were set to $a=20$ and $r=\frac{\pi}{2}$. We conducted stratified 5-fold cross-validation 10 times and calculated the AUC for the validation split at each epoch to select the best hyperparameters. Here, we used only the training data obtained from a random 9:1 split of the entire dataset, as described in Section 2.2. The test data were reserved for model evaluation and were not used during training or hyperparameter tuning. Stratification was based on normalized drug response values, divided into 4 equal-width bins between their minimum and maximum values. We took the average of all 50 runs and defined the best AUC value among 100 epochs as the performance of the model. Hyperparameters were determined by grid search (Table \ref{tab:tableA1}, Figure \ref{fig:figA1}, \ref{fig:figA2}, \ref{fig:figA3}, \ref{fig:figA4}, \ref{fig:figA5}). After completing the hyperparameter search, we evaluated the performance of the model using their best hyperparameters on the test set. The training data was split 4:1 into training and validation data using a stratified approach, as we did for the cross-validation. We trained models using the best hyperparameters for 100 epochs, calculating the AUC of the models on the validation set at the end of each epoch. For early stopping to prevent overfitting, we stopped training when the performance of the models on the validation set did not improve for three epochs, and used the models from the epoch just before the performance stopped improving as the final trained models. For all drugs, one of the quantum-classical hybrid models with the proposed method performed best. On the other hand, when normalization methods other than the proposed method were used, performance was worse than the classical model in some cases. It is suggested that if there is no appropriate normalization, the integration of quantum circuits can make the performance rather poor.
\begin{table}[H]
 \caption{Performance was evaluated on the randomly split test set. Proposed-multi means the proposed method with multiple measurements layer, and Proposed-single means that with a single measurement layer. The performance of the proposed methods is competitive or higher than other methods.}
  \centering
  \begin{tabular}{lllllll}
    \toprule
    Type & Normalization &   Cetuximab       &   Cisplatin       &   Docetaxel       &   Erlotinib       &   Gemcitabine  \\
    \midrule
    Classic  &               & $0.707$         &   $0.820$         &   $0.791$         &   $0.295$    &   $0.485$\\       
    Hybrid & Identity                &   $0.673$         &   $0.782$         &   $0.806$         &   $0.381$         &   $0.633$\\
    &Layernorm               &   $0.618$         &   $0.835$         &   $0.842$         &   $0.424$         &   $0.580$\\
    &$\tanh$                 &   $0.432$         &   $0.580$         &   $0.795$         &   $0.429$         &   $0.139$\\
    &Proposed-multi  &   $\bm{0.709}$    &   $\bm{0.827}$         &   $\bm{0.868}$    &   $0.443$         &   $0.679$\\
    &Proposed-single &   $0.707$         &   $0.707$    &   $\bm{0.868}$         &   $\bm{0.576}$         &   $\bm{0.685}$\\
    \bottomrule
  \end{tabular}
  \label{tab:table1}
\end{table}

\subsection{Hyperparameters of the normalization function}
To confirm the broad validity of this setup, we also investigated the effect of the hyperparameters $a$ and $r$ on the method’s performance. For each drug, we examined the change given different values of $a$ and $r$, for the best-performing model selected using cross-validation as described in section 3.1. As in 3.1, we trained the model using early stopping by splitting the training sample with a stratified approach and compared the performance on the test data. We fixed r to $\frac{\pi}{2}$ when changing a, and a to $20$ when changing r. Multiple measurements were taken from the circuit, which were combined using a single neural network layer.
For $a$, the performance was low when $a$ was small (0.5, 1) in most cases. On the other hand, the performance was high when $a$ took on values larger than 10. Gemcitabine is the only exception, and its performance is best at $a=0.5$.
For $r$, the optimal $r$ varies for each anticancer drug, suggesting that it is a parameter that needs to be adjusted. When $r=8\pi$, there was a substantial decrease in performance for drugs other than Erlotinib, suggesting that the periodicity issue of the quantum gate could not be ignored. For Erlotinib, while $r=8\pi$ recorded the highest performance, the overall performance remained low. This suggests that most models performed close to random chance, indicating that the prediction task for Erlotinib was too difficult.
\begin{table}[H]
 \caption{Performance with different $a$. $r$ was fixed to $\frac{\pi}{2}$.}
  \centering
  \begin{tabular}{llllll}
    \toprule
                       & Cetuximab & Cisplatin & Docetaxel & Erlotinib & Gemcitabine \\
    \midrule
    $a = 0.5$ & $0.577$ & $0.495$ & $0.539$ & $0.371$ & $\bm{0.790}$\\
    $a = 1$ & $0.432$ & $0.580$ & $0.795$ & $0.429$ & $0.139$\\
    $a = 10$ & $0.700$ & $0.716$ & $\bm{0.885}$ & $\bm{0.514}$ & $0.620$\\
    $a = 20$ & $0.709$ & $\bm{0.827}$ & $0.868$ & $0.443$ & $0.679$\\
    $a = 100$ & $\bm{0.734}$ & $0.727$ & $0.872$ & $0.448$ & $0.614$\\
    \bottomrule
  \end{tabular}
  \label{tab:table2}
\end{table}
\begin{table}[H]
 \caption{Performance with different $r$. $a$ was fixed to $20$. The optimal $r$ varies for each anticancer drug. 
 }
  \centering
  \begin{tabular}{llllll}
    \toprule
                       & Cetuximab & Cisplatin & Docetaxel & Erlotinib & Gemcitabine \\
    \midrule
    $r = \frac{\pi}{4}$  & $0.716$ & $0.782$ & $0.883$ & $0.386$ & $0.660$\\
    $r = \frac{\pi}{2}$  & $0.709$ & $\bm{0.827}$ & $0.868$ & $0.443$ & $0.679$\\
    $r = \frac{3\pi}{4}$ & $0.668$ & $0.730$ & $0.883$ & $0.476$ & $0.645$\\
    $r = \pi$            & $\bm{0.756}$ & $0.722$ & $\bm{0.885}$ & $0.324$ & $0.454$\\
    $r = \frac{3\pi}{2}$ & $0.742$ & $0.766$ & $0.857$ & $0.429$ & $0.596$\\
    $r = 2\pi$           & $0.711$ & $0.749$ & $0.816$ & $0.352$ & $0.620$\\
    $r = 4\pi$           & $0.675$ & $0.802$ & $0.855$ & $0.414$ & $\bm{0.685}$\\
    $r = 8\pi$           & $0.601$ & $0.584$ & $0.812$ & $\bm{0.538}$ & $0.404$\\
    \bottomrule
  \end{tabular}
  \label{tab:table3}
\end{table}

\subsection{Data normalization strategy}
\begin{figure}[H]
  \centering
  \includegraphics[width=0.6\columnwidth, trim={1mm 1mm 1mm 1mm}]{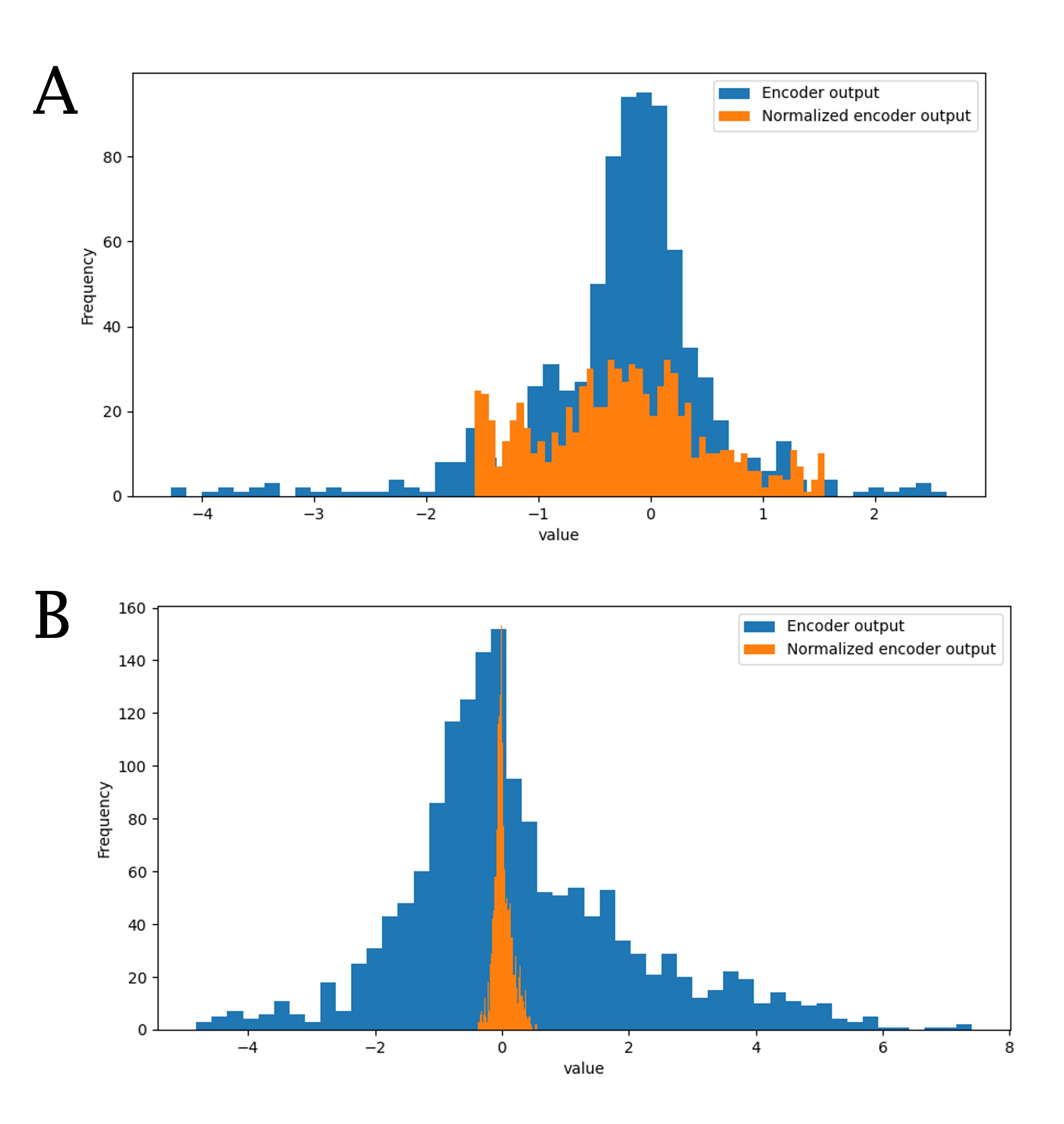}
  \caption{Distribution of values before and after normalization. The blue graph represents the features before normalization, and the orange graph represents the features after normalization. (A) Distribution of values before and after normalization of the model using the existing method described in Equation \ref{eq:eq1}. After normalization, values far from 0 are more densely distributed than before normalization. (B) Distribution of values before and after normalization of the model using the proposed method described in Equation \ref{eq:eq2}. ($a=20, r=\frac{\pi}{2}$). The shape of the original distribution remains preserved after normalization with a change in scale.}
  \label{fig:fig3}
\end{figure}

To understand how the proposed method normalizes the data, we examined the overall distribution of the transformed values. The following results were obtained using the predictive model of Docetaxel with the best hyperparameters trained in Section 3.1, due to its superior performance compared to the classical model. Specifically, we have looked at the values returned at the embedding layer in the conventional method, and after their transformation using tanh (Equation \ref{eq:eq1}), or the proposed method (Equation \ref{eq:eq2}). We extracted 100 training samples, input them into those trained models, and plotted how they were distributed before and after normalization (Figure \ref{fig:fig3}). In the conventional method, many values far from 0 are densely converted to values close to $-\frac{\pi}{2}$ or $\frac{\pi}{2}$, and the shape of the value distribution changes greatly from that before the normalization. On the other hand, the proposed method normalizes values keeping the shape of the original distribution.

\section{Discussion}
We found that combining quantum circuits with a classical model can deliver better performance in our application case of predicting anti-cancer drug response in cell line data for several drugs, but only when the proposed normalization function was used. We believe this to be the case because this transformation prevents extreme values from clustering densely after normalization. In this paper, we were only able to validate the performance using randomly split test data rather than independent external test data. As this is an important limitation, we would like to conduct such additional tests using real patient data in the future.\\
We only compared our model with a simple neural network model, excluding quantum circuits, and do not claim a general performance advantage over classical models in the field of anti-cancer drug response prediction. We will consider more advanced quantum-classical hybrid anti-cancer drug response prediction models using the normalization method proposed.\\
We found that small $a$ and large $r$ are not well suited for the proposed normalization function for most prediction models. The former is due to value crowding, and the latter due to the periodicity property of quantum gates. However, there are exceptions, such as the prediction model for Gemcitabine, which prefers a small $a$, and the prediction model for Erlotinib, which prefers a large $r$. The optimal values of $a$ and $r$ differ for each prediction model, and this parameter should be determined carefully. One potential method of determining $a$ and $r$ is to set $a$ and $r$ as parameters and adjust them during the training process to search for the optimal values. In future work, we plan to conduct experiments using noisy simulators and real quantum devices to investigate the effect of noise on the optimal hyperparameters a and r. This will be a crucial step towards understanding the performance of our method in realistic quantum devices.\\
The applicability of the proposed normalization method to fields other than the prediction of response to anti-cancer drugs is one of the themes that should be studied in the future. In particular, the similar model architecture with normalization may be applicable to fields such as image recognition, where high-dimensional data is handled and the input method to quantum circuits is not obvious.
\section{Conclusion}
In this study, we found that appropriate normalization improves the performance of the quantum-classical hybrid model in the prediction of anti-cancer drug response using gene expression levels. The proposed normalization of the method enables stable learning by avoiding the periodicity problem of quantum gates and the crowding of values caused by the existing normalization methods. Quantum computers have developed remarkably in recent years, and their application in the field of biomedical science is expected to become more important in the future. Further progress in omics data analysis is expected through research that combines deep learning and quantum computers using this method.

\section*{Acknowledgments}
We would like to thank Drs. Mitsuhisa Sato, Maxence Vandromme, Miwako Tsuji, Kengo Nakajima, Yuetsu Kodama, and Kazuya Yamazaki for their helpful discussions. This study was supported by the World-leading Innovative Graduate Study Program in Proactive Environmental Studies (WINGS-PES), The University of Tokyo. This work was also partly funded by JSPS KAKENHI Grant Numbers JP25K02261, Japan. 

\bibliographystyle{unsrt}  
\bibliography{optimalnorm}

\begin{thebibliography}{10}

\bibitem{sharifi2019moli}
Hossein Sharifi-Noghabi, Olga Zolotareva, Colin~C Collins, and Martin Ester.
\newblock Moli: multi-omics late integration with deep neural networks for drug response prediction.
\newblock {\em Bioinformatics}, 35(14):i501--i509, 2019.

\bibitem{sharma2023deepinsight}
Alok Sharma, Artem Lysenko, Keith~A Boroevich, and Tatsuhiko Tsunoda.
\newblock Deepinsight-3d architecture for anti-cancer drug response prediction with deep-learning on multi-omics.
\newblock {\em Scientific reports}, 13(1):2483, 2023.

\bibitem{xie2020smooth}
Cihang Xie, Mingxing Tan, Boqing Gong, Alan Yuille, and Quoc~V Le.
\newblock Smooth adversarial training.
\newblock {\em arXiv preprint arXiv:2006.14536}, 2020.

\bibitem{geeleher2014clinical}
Paul Geeleher, Nancy~J Cox, and R~Stephanie Huang.
\newblock Clinical drug response can be predicted using baseline gene expression levels and in vitro drug sensitivity in cell lines.
\newblock {\em Genome biology}, 15:1--12, 2014.

\bibitem{senokosov2024quantum}
Arsenii Senokosov, Alexandr Sedykh, Asel Sagingalieva, Basil Kyriacou, and Alexey Melnikov.
\newblock Quantum machine learning for image classification.
\newblock {\em Machine Learning: Science and Technology}, 5(1):015040, 2024.

\bibitem{paszke2019pytorch}
A~Paszke.
\newblock Pytorch: An imperative style, high-performance deep learning library.
\newblock {\em arXiv preprint arXiv:1912.01703}, 2019.

\bibitem{bergholm2018pennylane}
Ville Bergholm, Josh Izaac, Maria Schuld, Christian Gogolin, Shahnawaz Ahmed, Vishnu Ajith, M~Sohaib Alam, Guillermo Alonso-Linaje, B~AkashNarayanan, Ali Asadi, et~al.
\newblock Pennylane: Automatic differentiation of hybrid quantum-classical computations.
\newblock {\em arXiv preprint arXiv:1811.04968}, 2018.

\bibitem{hendrycks2016gaussian}
Dan Hendrycks and Kevin Gimpel.
\newblock Gaussian error linear units (gelus).
\newblock {\em arXiv preprint arXiv:1606.08415}, 2016.

\bibitem{mari2020transfer}
Andrea Mari, Thomas~R Bromley, Josh Izaac, Maria Schuld, and Nathan Killoran.
\newblock Transfer learning in hybrid classical-quantum neural networks.
\newblock {\em Quantum}, 4:340, 2020.

\bibitem{glorot2010understanding}
Xavier Glorot and Yoshua Bengio.
\newblock Understanding the difficulty of training deep feedforward neural networks.
\newblock In {\em Proceedings of the thirteenth international conference on artificial intelligence and statistics}, pages 249--256. JMLR Workshop and Conference Proceedings, 2010.

\bibitem{sagingalieva2023hybrid}
Asel Sagingalieva, Mohammad Kordzanganeh, Nurbolat Kenbayev, Daria Kosichkina, Tatiana Tomashuk, and Alexey Melnikov.
\newblock Hybrid quantum neural network for drug response prediction.
\newblock {\em Cancers}, 15(10):2705, 2023.

\bibitem{perez2020data}
Adri{\'a}n P{\'e}rez-Salinas, Alba Cervera-Lierta, Elies Gil-Fuster, and Jos{\'e}~I Latorre.
\newblock Data re-uploading for a universal quantum classifier.
\newblock {\em Quantum}, 4:226, 2020.

\bibitem{schuld2021effect}
Maria Schuld, Ryan Sweke, and Johannes~Jakob Meyer.
\newblock Effect of data encoding on the expressive power of variational quantum-machine-learning models.
\newblock {\em Physical Review A}, 103(3):032430, 2021.

\bibitem{kingma2014adam}
Diederik~P Kingma and Jimmy Ba.
\newblock Adam: A method for stochastic optimization.
\newblock {\em arXiv preprint arXiv:1412.6980}, 2014.

\bibitem{yang2012genomics}
Wanjuan Yang, Jorge Soares, Patricia Greninger, Elena~J Edelman, Howard Lightfoot, Simon Forbes, Nidhi Bindal, Dave Beare, James~A Smith, I~Richard Thompson, et~al.
\newblock Genomics of drug sensitivity in cancer (gdsc): a resource for therapeutic biomarker discovery in cancer cells.
\newblock {\em Nucleic acids research}, 41(D1):D955--D961, 2012.

\bibitem{alam2021quantum}
Mahabubul Alam, Satwik Kundu, Rasit~Onur Topaloglu, and Swaroop Ghosh.
\newblock Quantum-classical hybrid machine learning for image classification (iccad special session paper).
\newblock In {\em 2021 IEEE/ACM International Conference On Computer Aided Design (ICCAD)}, pages 1--7. IEEE, 2021.

\end{thebibliography}

\clearpage
\section*{Appendix}
\renewcommand{\thetable}{A\arabic{table}}
\renewcommand{\thefigure}{A\arabic{figure}}
\setcounter{table}{0}
\setcounter{figure}{0}
We determined the optimal parameters by cross-validation. We performed grid searches over the range in Table A1 for the hyperparameters and selected the combination that recorded the highest average AUC. We plotted the prediction performance of each model by epoch (Figure A1, A2, A3, A4, A5), showing the average of 50 runs of AUC for the validation data obtained from 10 times 5-fold cross-validation. The red dots indicate the points with the highest AUC.
\begin{table}[h]
 \caption{Grid search was conducted with this set of hyperparameters.}
  \centering
  \begin{tabular}{lc}
    \toprule
    Hyperparameter                   &   Range       \\
    \midrule
    Number of qubits ($n_1$) & $[4, 8]$  \\
    Number of encoding layers ($n_2$) & $[2, 4]$ \\
    Number of variational layers ($n_3$) & $[1, 2, 4]$ \\
    learning rate & $[1e-6, 1e-5, 1e-4]$\\
    \bottomrule
  \end{tabular}
  \label{tab:tableA1}
\end{table}
\clearpage
\begin{figure}
  \centering
  \begin{minipage}[b]{0.49\columnwidth}
      \includegraphics[width=0.9\columnwidth]{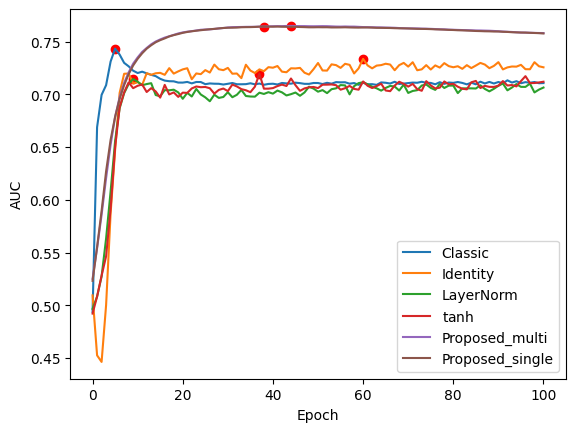}
      \caption{Cetuximab}
      \label{fig:figA1}
  \end{minipage}
  \begin{minipage}[b]{0.49\columnwidth}
      \includegraphics[width=0.9\columnwidth]{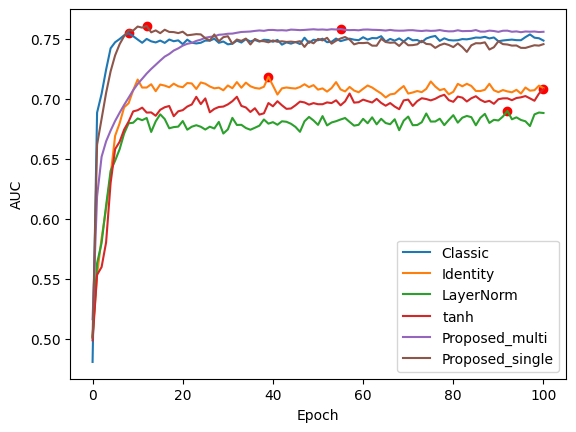}
      \caption{Cisplatin}
      \label{fig:figA2}
  \end{minipage}
\end{figure}
\begin{figure}
  \centering
  \begin{minipage}[b]{0.49\columnwidth}
      \includegraphics[width=0.9\columnwidth]{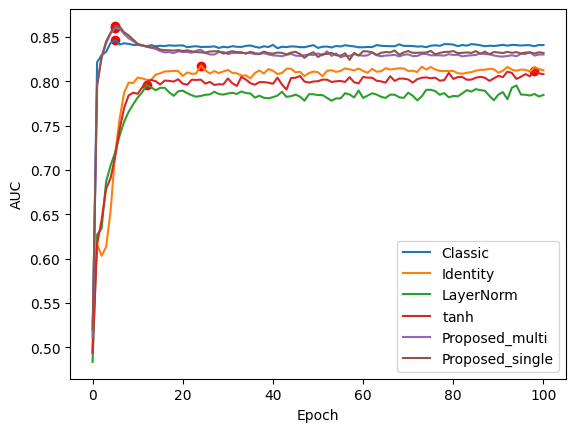}
      \caption{Docetaxel}
      \label{fig:figA3}
  \end{minipage}
  \begin{minipage}[b]{0.49\columnwidth}
      \includegraphics[width=0.9\columnwidth]{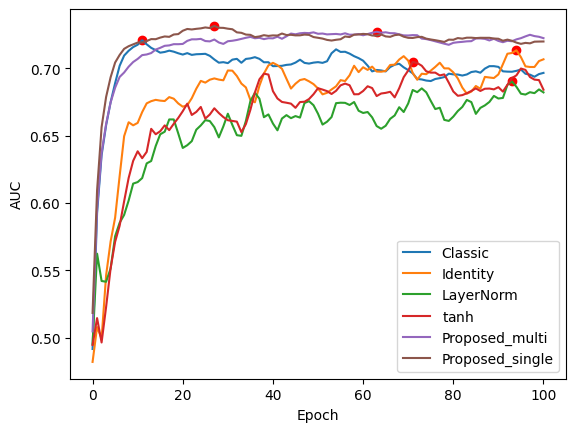}
      \caption{Erlotinib}
      \label{fig:figA4}
  \end{minipage}
\end{figure}
\begin{figure}
  \centering
  \begin{minipage}[b]{0.49\columnwidth}
      \includegraphics[width=0.9\columnwidth]{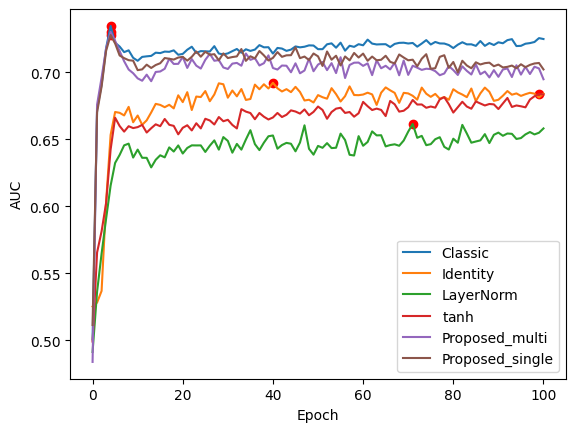}
      \caption{Gemcitabine}
      \label{fig:figA5}
  \end{minipage}
\end{figure}

\end{document}